\documentclass[twoside,11pt]{article}

\usepackage{jmlr2eArxiv}
\usepackage{amsmath}    
\usepackage{color}  
\hypersetup{
    colorlinks=true,
    citecolor=blue,
}

\providecommand{\ceil}[1]{\left\lceil{#1}\right\rceil}

\def\D{{\cal D}}

\def\R{{\mathbb R}}

\def\P{{\cal P}}
\def\F{{\cal F}}
\def\X{{\cal X}}
\def\Y{{\cal Y}}
\def\bX{{\bf X}}
\def\PROB {{\mathbb P}}
\def\EXP {{\mathbb E}}
\def\IND{{\mathbb I}}

\def \eD {\stackrel{{\cal D}}{=}}

\def\argmin{\mathop{\rm arg\, min}}
\def\argmax{\mathop{\rm arg\, max}}
\def\dist{\nu}
\def\marg{\mu}
\newcommand{\nrm}[1]{\left\Vert #1 \right\Vert}
\newcommand{\eps}{\varepsilon}
\newcommand{\supp}{\operatorname{supp}}

\newcommand{\dP}{R}

\newcommand{\gproto}{\tilde g}
\newcommand{\gkproto}{\hat{g}}

\newcommand{\proto}{\text{Proto-NN}}
\newcommand{\optinet}{\text{OptiNet}}
\newcommand{\hybrid}{\text{Proto-$k$-NN}}

\newcommand{\ft}{f^t}
\newcommand{\G}{\mathcal G}

\ShortHeadings{Universal consistency and convergence rates for prototype algorithms}{Gy\"orfi and Weiss}
\firstpageno{1}

\begin{document}

\title{Universal consistency and rates of convergence of multiclass prototype 
algorithms
 in metric spaces
}

\author{\name L\'aszl\'o Gy\"orfi \email gyorfi@cs.bme.hu \\
       \addr Department of Computer Science and Information Theory\\
       Budapest University of Technology and Economics\\
       1111 Budapest, Hungary
       \AND
       \name Roi Weiss \email roiw@ariel.ac.il \\
       \addr Department of Computer Science\\
       Ariel University\\
       40800 Ariel, Israel}

\editor{Samory Kpotufe}

\maketitle

\begin{abstract}
We study universal consistency and 
convergence 
rates
of simple nearest-neighbor prototype rules for the problem of multiclass classification in metric spaces. 
We first show that a novel data-dependent partitioning rule,
named \proto{}, is universally
consistent in any 
metric space that admits a universally consistent rule.
\proto{} is a significant simplification of \optinet{}, a recently proposed compression-based algorithm that, 
to date,
was 
the only algorithm known to be universally consistent in such a general setting.
Practically, \proto{} is simpler to implement and enjoys reduced computational complexity.

We then 
proceed to 
study convergence rates of the excess error probability.
We first obtain 
rates for the standard $k$-NN rule under a margin condition and a new generalized-Lipschitz condition.
The latter is an extension of a recently proposed modified-Lipschitz condition from $\R^d$ to metric spaces. Similarly to the modified-Lipschitz condition, the new condition avoids any boundness assumptions on the 
data distribution.
While obtaining rates for 
\proto{} 
is left open,
we show that a second prototype rule that hybridizes between $k$-NN and \proto{}
achieves the same rates as $k$-NN while enjoying similar computational advantages as \proto{}.
\textcolor{black}{However, as $k$-NN, 
this hybrid rule
is not consistent in general.}
\end{abstract}

\smallskip


\section{Introduction}
\label{sec:intro}
Let $(\X,\rho)$ be a separable metric space, equipped with its
Borel $\sigma$-field
\citep{CoHa67}.
Assume that the feature element $X$ takes values in $\X$ and let its label $Y$ take values in $\Y=\{1,\dots ,M\}$.
The error probability of an arbitrary decision function $g:\X\to\Y$ is
\[
L(g)=\PROB\{g(X)\ne Y\}.
\]
Denote by $\dist$ the unknown probability distribution
of $(X,Y)$
and let
\[
P_j(x)=\PROB\{Y=j\mid X=x\},\quad
j\in\Y.
\]
Then the Bayes decision,
\[
g^*(x) = \argmax_{j\in\Y}  P_j(x),
\]
minimizes the error probability.
This optimal error is
denoted by
\[
L^*=\PROB\{g^*(X)\ne Y\}.
\]

In the standard model of pattern recognition, we are given
labeled samples,
$\D_n=\{(X_1,Y_1), \dots, (X_n,Y_n)\},$
which are $n$ independent copies of $(X,Y)$.
Based on 
$\D_n$,
one can estimate the regression
functions $P_j$ by $P_{j,n}$,
$j\in\Y$,
and the 
plug-in classification rule $g_n$ 
derived from $P_{j,n}$ 
is
\[
g_n(x) = \argmax_{j\in\Y} P_{j,n}(x) .
\]
The classifier $g_n$ is
\emph{weakly consistent} for
the distribution
$\dist$
if
\begin{align*}
\lim_{n\to\infty} \EXP\{L(g_{n})\} = L^*.
\end{align*}
It is
\emph{strongly consistent} for $\dist$ if
\begin{align*}
\PROB\{\lim_{n\to\infty} L(g_{n}) = L^* \} = 1.
\end{align*}
The classifier $g_n$ is
\emph{universally consistent} in the metric space $(\X,\rho)$ if it is consistent for \emph{any} distribution $\dist$ over the Borel $\sigma$-field.

Following the pioneering work of \citet{CoHa67} and \citet{stone1977}
on nearest-neighbor classification,
%
%
\citet{MR877849,DeGyLu96,GyKoKrWa02} showed that the $k$-nearest neighbor rule ($k$-NN) is universally strongly consistent in the Euclidean space
$(\R^d,\nrm{\cdot}_2)$
provided that $k\to\infty$ and $k/n\to0$ as $n\to\infty$.
Kernel-based and various partitioning rules were shown to be universally consistent in $\R^d$ as well \citep{DeGyLu96}.
%
%
For more general (and, in particular, infinite-dimensional) metric spaces, \citet{cerou2006nearest,forzani2012consistent}
characterized the consistency of the $k$-NN rule
in terms of a Besicovitch-type condition.
In particular, \citet{cerou2006nearest} showed that in the binary case, $|\Y|=2$, a
\textcolor{black}{sufficient} condition for the $k$-NN classifier to be weakly consistent for $\dist$ is
\begin{equation}
\label{eq:besic_cerou}
\lim_{r\to 0^+}
\PROB
\left\{
\frac{1}{\marg(S_{X,r})} \int_{S_{X,r}} |P_j(z) - P_j(X)| \marg(dz)
> \eps
\right\}
= 0.
\end{equation}
Here, $S_{x,r}=\{z\in\X: \rho(x,z)\leq r\}$ denotes the closed ball centered at $x$ and having radius $r$ and $\marg$ is the marginal distribution of $X$.
\textcolor{black}{It was also shown in \cite{cerou2006nearest} that in the realizable case, where $P_1(x)\in\{0,1\}$ for all $x\in X$, a violation of \eqref{eq:besic_cerou} implies that $k$-NN is \emph{inconsistent} \textcolor{black}{(see also \citet{cesari2021nearest})}.}
\citet{MR2327897} established the strong consistency of a generalized moving-window rule under the same condition \eqref{eq:besic_cerou}.

By Besicovitch's density theorem
\citep{Fe69},
in
$\R^d$
 --- and more generally in any {finite-dimensional} normed space ---
condition \eqref{eq:besic_cerou} holds for \emph{all} distributions $\dist$.
However, in infinite-dimensional spaces this condition may be violated \citep{preiss1979invalid,MR609946}.
As such, the $k$-NN and the moving-window classifiers are not universally consistent in general separable metric spaces.
This violation
is not an isolated pathology, occurring, for example, in the commonly used separable Gaussian Hilbert spaces \citep{MR1974687}.
%
Leveraging the consistency of $k$-NN in finite dimensions, the {filtering} technique (i.e., taking the first $d_n$ coordinates in some basis representation for an appropriate $d_n$) was shown to be universally weakly consistent by \citet{BiBuWe05}.
However, that technique is only applicable in Hilbert spaces, as opposed to more general metric spaces.

Up until recently, sufficient and necessary conditions on a metric space under which a universally consistent rule exists in it was an open problem.
Building upon the results of \citet{kontorovich2014bayes,KoSaWe17}
it has been recently shown by \citet{HaKoSaWe20} that
a compression-based prototype classification rule named \optinet{} is universally strongly consistent in \emph{any} separable metric space 
\textcolor{black}{(for example, the space $L^p([a,b])$ is a metric space for any $p\geq 1$ and is separable for $1\leq p < \infty$ and non-separable for $p=\infty$)}.
Moreover,
\optinet{} was shown to be universally strongly consistent in any metric space that admits a universally consistent classifier --- the first algorithm known to enjoy this property.

Arguably, \optinet{} is a rather complex algorithm (see Section~\ref{sec:univ}).
The first main result of this paper (Corollary~\ref{thm0}) distills the core arguments used to establish \optinet{}'s consistency and shows that
a novel, much simpler classification rule
is universally strongly consistent in any metric space that admits a universally consistent classifier.
In addition, its consistency proof is much simpler than that of \optinet{} and its implementation is trivial.
As \optinet{}, this algorithm, henceforth termed \proto{} (and formally presented in Section~\ref{sec:univ}), is a prototype algorithm and has the advantage of reduced computational complexity.
Concerning some other prototype nearest-neighbor rules, see 
 \citet[\S19.3]{DeGyLu96}.

Another important property of a classification rule is the rate at which the excess error probability $\EXP\{L(g_{n})\} - L^*$ convergences to 0
as $n\to\infty$.
For any classification rule $g_n$, this rate
can be arbitrarily slow without further assumptions on the
unknown
data distribution \citep{DeGyLu96}.
So to obtain a non-trivial rate of convergence
one typically assumes that $\dist$ belongs to a large class of distributions meeting some smoothness and tail conditions.

For the case $\X=\R^d$, rates of convergence for several algorithms were obtained under a variety of conditions \citep{DeGyLu96,GyKoKrWa02}.
A common assumption is that the regression functions are
H\"{o}lder-continuous, that is, there are some $C>0$ and $0<\beta\leq 1$ such that $\forall x,z\in\X$,
\begin{align}
\label{eq:Lip}
|P_j(x)-P_j(z)| \leq C \rho(x,z)^{\beta}.
\end{align}
The case $\beta=1$ is know as Lipschitz continuity.
Denoting the support of $\marg$ by 
$$\supp(\marg)=\{x\in\X:\marg(S_{x,r})>0,\forall r>0\},$$
it is well known \citep{GyKoKrWa02}
that 
if $\supp(\marg)$ is bounded and the regression function
is H\"{o}lder-continuous,
then, for
$|\Y|=2$,
the $k$-NN rule
with $k=n^{\frac{2\beta}{2\beta+d}}$
achieves
the rate
\begin{align*}
\EXP\{L(g_{k,n})\} - L^* = 
O( n^{-\frac{\beta}{2\beta + d}}).
\end{align*}
While this rate holds also for the more general problem of $L^1$ real-valued bounded regression, \citet{mammen1999smooth,tsybakov2004optimal,AuTs05} showed that for binary classification, faster rates can be achieved under an additional margin condition.
Recently, 
\citet{xue2018achieving,puchkin2020adaptive} generalized the margin condition to 
the 
multiclass
setting.

\begin{definition}
\label{def:marg}
Let $P_{(1)}(x)\ge \dots \ge P_{(M)}(x)$ be the ordered values of
$P_{1}(x), \dots , P_{M}(x)$.
Then the {\bf \textit{margin condition}}  means that
there are some $\alpha>0$ and $c^*>0$
such that
\begin{equation}
\label{wtsyb}
\PROB\{P_{(1)}(X)-P_{(2)}(X)\le t\}
\le c^* t^{\alpha}
,\qquad 0 < t \leq 1.
\end{equation}
\end{definition}

Assuming that the regression function is H\"older-continuous, that the margin condition is satisfied, 
and that the marginal distribution $\marg$ of $X$ has a density 
that satisfies the so called strong density condition, \citet{AuTs05,KoKr07, GaKlMa16} showed that
for
the binary case,
the $k$-NN rule and several other plug-in estimators
achieve the rate
\begin{align}
\label{eq:rate_Rd}
\EXP\{L(g_{k,n})\} - L^* = O( n^{-\frac{\beta(1+\alpha)}{2\beta + d}}).
\end{align}
Moreover, this rate was shown to be minimax optimal, meaning that, over the class of all distributions meeting the aforementioned conditions, this rate is also a lower bound 
for \emph{any} classification rule.
See \citet{samworth2012,blaschzyk2018improved} and references therein for even faster rates under stronger conditions on the regression function.

The strong density condition
under which
the rates in \eqref{eq:rate_Rd} are established requires the density function 
to be bounded away from zero over the support of $\marg$;
a highly restrictive condition that does not hold for many distributions of practical interest.
Recently, it has been shown by \citet{DoGyWa18} that the same rates  in \eqref{eq:rate_Rd} are obtained by replacing the strong density condition, together with the H\"older condition \eqref{eq:Lip}, with
a combined smoothness and tail condition, named modified Lipschitz condition, given by
\begin{equation}
\label{eq:mod-Lip}
|P_j(x)-P_j(z)|\le C\marg(S_{x,\rho(x,z)})^{\beta/d}.
\end{equation}
\citet{ChDa14} considered the related condition that there are
 some $\gamma>0$ and $C^*>0$ such that for all $x$ in the support of $\marg$,
\begin{align}
\label{eq:dasg_cond}
\left| P_j(x) - \frac{1}{\marg(S_{x,r})}{\int_{S_{x,r}} P_j(z) d\marg(z)} \right|
\leq C^* \marg(S^o_{x,r})^\gamma.
\end{align}
Here, $S^o_{x,r}=\{z\in\X: \rho(x,z) < r\}$ denotes the open ball centered at $x$.
They showed
that
in the binary case,
under condition \eqref{eq:dasg_cond} and the margin condition \eqref{wtsyb},
the $k$-NN rule
achieves
\begin{align}
\label{eq:dasg_rate}
\EXP\{L(g_{k,n})\} - L^* = O( n^{-\frac{\gamma(1+\alpha)}{2\gamma + 1}}).
\end{align}
Evidently, for $\X=\R^d$, the rates in \eqref{eq:rate_Rd} are revisited by setting $\gamma=\beta/d$.
More generally,
condition \eqref{eq:dasg_cond} can be seen as a uniform Besicovitch condition and is a-priori applicable in any separable metric space.
%
In this paper, we further abstract the modified Lipschitz conditions \eqref{eq:mod-Lip} and \eqref{eq:dasg_cond} and consider the following combined smoothness and tail condition.
\begin{definition}
\label{def:genLip}
For each $j\in\Y$, the function $P_j$
satisfies the {\bf \textit{generalized Lipschitz condition}} if there is a monotonically increasing function $h: [0,1]\to \R^+$ with $h(s)\downarrow 0$ as  $s\downarrow 0$ such that for any
 $x,z\in \X$,
\begin{equation}
\label{gen-Lip}
|P_j(x)-P_j(z)|\le h(\marg(S_{x,\rho(x,z)})).
\end{equation}
\end{definition}

\medskip

In Section~\ref{sec:rates}, we first obtain rates for the $k$-NN rule under the generalized Lipschitz condition and the margin condition in terms of the function $h$ (Theorem~\ref{thm:knn_rate}).
For the case
\begin{align}
\label{eq:h_gamma}
h(s) = C^* s^\gamma
\end{align}
we revisit the same rates as in \eqref{eq:dasg_rate}.
While obtaining rates for \proto{} is left open,
we proceed to derive rates for a second novel prototype rule that hybridizes between \proto{} and $k$-NN 
(Theorem~\ref{thm:kproto_rate}).
This rule, which we call \hybrid{}, allows a reduction in computational complexity by compressing the data into $m=O(n/k)$ prototypes while enjoying the same rates as $k$-NN;
see also \citet{xue2018achieving} for results in the same spirit.

\medskip

\paragraph{On the generalized Lipschitz condition.} Note that for condition \eqref{gen-Lip} to be non-trivial, the rate at which $h(s)\to 0$ as $s\to 0$ should appropriately reflect the geometry of $\X$ and the class of distributions $\marg$ under consideration.
Consider for example
two distinct points
$x,z\in\R^d$
lying in a region where $\marg$ is uniformly distributed.
Denoting the Lebesgue measure by $\lambda$ and
abbreviating its measure on any ball
by $\lambda(S_{x,r}) = v_d \cdot r^d$,
one can verify that for any $0< r \ll \rho(x,z)$, applying the triangle inequality for
$\ceil{\rho(x,z)/r}$
times  over a path from $x$ to $z$, the generalized Lipschitz condition \eqref{gen-Lip} implies
that for some constant $c>0$,
\begin{align*}
|P_j(x) - P_j(z)| &\le
c
\cdot
\left(\frac{\rho(x,z)}{r}\right)
\cdot
h(\lambda(S_{x,r}))
 = \rho(x,z) \cdot O\left( \frac{h(r^d)}{r} \right)
\qquad \text{as } r\to 0.
\end{align*}
Hence, to allow for non-constant regression functions,
one needs
$h(\lambda(S_{x,r}))=O(r)$, or equivalently $h(s) = O(s^{1/d})$, as in the modified Lipschitz condition \eqref{eq:mod-Lip}.
More generally, assuming that $\marg$ is absolutely continuous with respect to
$\lambda$,
the Radon-Nikodym theorem
\citep{Fe69}
asserts that $\marg$ has a density, namely, there exists a function $D:\R^d\to\R^+$ such that for $\lambda$-almost all $x\in\R^d$,
\begin{align}
\label{eq:diff_thm}
\lim_{r\to 0} \left|\frac{\marg(S_{x,r})}{\lambda(S_{x,r})}- D(x)\right| = 0,
\end{align}
such that for any measurable $A\subseteq\R^d$,
$$\marg(A)=\int_{A} D(x)\lambda(dx).$$
So, in this case, condition \eqref{gen-Lip} 
with
\begin{align}
\label{eq:h_Rd}
h(s) = C^* s^{\beta/d}
\end{align}
essentially becomes
\begin{equation*}
|P_j(x)-P_j(z)|\le C^* D(x)^{\beta/d} \rho(x,z)^\beta.
\end{equation*}
This condition should be compared to the  H\"older condition \eqref{eq:Lip};
see also \citet{gyorfi1981rate}.
Similarly, for a general separable metric space, 
one may assume that $\marg$
is accompanied by
an increasing ``small-ball probability'' function $\psi:\R^+ \to \R^+$ with $\lim_{r\to0} \psi(r)=0$ and a function $K:\X\to\R^+$ such that for
$\marg$-almost all $x\in\X$,
\begin{equation}
\label{eq:ball_psi}
\marg(S_{x,r}) \leq K(x)\psi(r)
\qquad \text{as } r\to0.
\end{equation}
In this case, an appropriate choice would be
$$
h(s)= C^*\psi^{-1}(s)^\beta
\qquad s\in[0,1].
$$
Condition \eqref{gen-Lip} then becomes
\begin{equation*}
|P_j(x)-P_j(z)|\le C^* \psi^{-1}\big(K(x)\psi(\rho(x,z))\big)^\beta.
\end{equation*}

As an example, consider a doubling measure $\marg_0$  with $\supp(\marg_0)=\X$
\citep{heinonen2012lectures,rigot2018differentiation}.
Such a measure satisfies that there exists $C = C(\marg_0)\geq1$ such that for any $x$
 and any radius $r > 0$,
\begin{equation}
\label{eq:doubling}
0 < \marg_0(S_{x,2r}) \leq C \marg_0(S_{x,r}).
\end{equation}
Assuming that $\mu_0$ has no atoms,
the doubling property \eqref{eq:doubling} implies
that there exists $\tau>0$ such that
\begin{equation}
\label{eq:psi_finite}
\marg_0(S_{x,r}) =
\Theta(\psi(r))=
\Theta(r^\tau) \qquad \text{as }r\to0.
\end{equation}
Since doubling measures satisfy a differentiation theorem similar to \eqref{eq:diff_thm},
a large class of distributions that satisfy
\eqref{eq:ball_psi}
consists of all distributions $\marg$ that are absolutely continuous with respect to $\marg_0$.
Here, $K$ in \eqref{eq:ball_psi} is again related to $\marg$'s density
with respect to
$\marg_0$.
It is worth mentioning,
however, that in infinite-dimensional spaces, doubling measures may not exist and a differentiation theorem 
may not hold; see
\citet{heinonen2012lectures,rigot2018differentiation}.

In the field of non-parametric functional data analysis \citep{ferraty2006nonparametric,burba2009k,baillo2011classification,ling2018nonparametric},
where input data items
are in the form of random functions, distributions $\marg$ accompanied by $\psi$ in the form of \eqref{eq:psi_finite} are said to have a fractal dimension $\tau$;
see \citet{pesin1993rigorous, bardet1997tests} for concrete examples.
Similarly to \eqref{eq:dasg_rate}, a finite fractal dimension leads to rates of order $n^{-\xi}$ for an appropriate $\xi=\xi(\tau,\alpha,\beta)>0.$
However, in infinite-dimensional spaces,
one typically encounters
distributions of exponential type, where for some $C>0$ and $\tau,\tau'>0$,
\begin{equation}
\label{eq:psi_exp}
\psi(r) = \Theta\big( e^{-\frac{1}{r^{\tau}}\log\left(\frac{1}{r}\right)^{\tau'}}\big)
\qquad \text{as } r\to 0.
\end{equation}
Such distributions include, for example, various diffusion and Gaussian processes; see
\citet[\S13]{ferraty2006nonparametric} and references therein.
In this case, non-parametric estimators suffer from extremely slow convergence rates of order $\log(n)^{-\xi}$,
even for estimating the regression function at a \emph{fixed} point $x\in\X$ \citep{ferraty2006nonparametric, mas2012lower}.
One may overcome such slow rates by considering a \textcolor{black}{pseudo}-metric over $\X$ (where the condition $\rho(x,y)=0\Rightarrow x=y$ is removed)
 instead of a metric, 
 but coming up with an appropriate \textcolor{black}{pseudo}-metric can be a challenging task \citep{ferraty2006nonparametric}.

Lastly, 
we note that
faster rates
can be achieved in infinite-dimensional spaces, uniformly over the support of $\marg$, in terms of covering numbers, assuming the support
of $\marg$ is totally bounded
\citep{kulkarni1995rates,ferraty2010rate,kudraszow2013uniform,BiCeGu10}.
Here we do not make such an assumption.


\section{Universal consistency of
\proto{}
 }
\label{sec:univ}
In this section we first show that a simple prototype
classification rule, which we call \proto{}, is universally strongly consistent in any
separable metric space.
By the recent results of \citet{HaKoSaWe20}, this implies that \proto{} is in fact universally strongly consistent in any metric space that admits a universally consistent rule; see also \citet{collins2020universal}.
\textcolor{black}{Our consistency result is obtained by first establishing the consistency of \proto{} for the more general problem of real-valued bounded $L^1$-regression. Lastly, we show that a slightly modified version of \proto{} is universally strongly consistent for the general real-valued $L^p$-regression problem for any $1\leq p < \infty$ under the (necessary) condition that $\EXP\{|Y|^p\}<\infty$.}

To simplify 
\proto{}'s
analysis, we assume that in addition to the labeled sample 
$\D_n$,
we also have an independent unlabeled sample $\bX'_m =\{X'_1,\dots ,X'_m\}$ 
where the
$X'_i$'s are independent copies of $X$.
Introduce the data-driven partition $\P_{m}$ of $\X$ such that $\P_{m}$ is a Voronoi partition with the nucleus set $\bX'_m$, i.e., 
$$\P_{m}=\{A_{m,1},A_{m,2},\dots ,A_{m,m}\}$$
such that $A_{m,\ell}$ is the Voronoi cell around the nucleus $X'_\ell$,
\[
A_{m,\ell} = \left\{x\in\X : \ell = \argmin_{1\leq i \leq m} \rho(x,X'_i)\right\},
\]
where tie breaking is done by indices, i.e., if $X'_i$ and $X'_j$ are equidistant
from $x$, then $X'_i$ is declared ``closer'' if $i < j$.
\proto{} estimates the regression function $P_j$ over each cell by
the piecewise constant function
\begin{align}
\label{eq:proto_est}
\tilde P_{n,j}(x)=\frac{\sum_{i=1}^n \IND_{\{Y_i=j,X_i\in A_{m,\ell}\}}}{\sum_{i=1}^n \IND_{\{X_i\in A_{m,\ell}\}}},\quad \mbox{ if } x\in A_{m,\ell},
\end{align}
such that $0/0=0$ by definition.
\proto{}
is then defined by
\begin{align}
\label{eq:gproto}
\gproto_{n}(x)=\argmax_{j\in\Y} \tilde P_{n,j}(x).
\end{align}
This rule is just the empirical majority vote over the labeled samples from $\D_n$ that fell into the cell in which $x$ resides, as determined by $\bX'_m$.
\textcolor{black}{
\proto{}'s construction time 
is $O(m n)$ and a query takes $O(m)$ time.
}

%
%
%

To establish \proto{}'s universal consistency, we first establish its consistency for the more general problem of $L^1$ real-valued bounded regression
\citep{GyKoKrWa02}.
Formally, let the label $Y$ be real-valued, and denote the corresponding regression function by
\begin{align*}
f(x)=\EXP\{Y\mid X=x\}.
\end{align*}
Introduce the partitioning regression estimate:
\begin{align}
\label{eq:reg_est}
f_{n}(x)=\frac{\sum_{i=1}^n Y_i\IND_{\{X_i\in A_{m,\ell}\}}}{\sum_{i=1}^n \IND_{\{X_i\in A_{m,\ell}\}}}, \quad \mbox{ if } x\in A_{m,\ell}.
\end{align}
\begin{theorem}
\label{thm2}
\textcolor{black}{Let $(\X,\rho)$ be a separable metric space.}
If $Y$ is bounded and $m=m_n\to \infty$ such that $m_n/n\to 0$,
then the 
estimate $f_{n}$ is strongly consistent in $L^1$, that is, for any distribution $\dist$ of $(X,Y)$,
\begin{align*}
\lim_{n\to\infty} \int |f_n(x)-f(x)|\marg(dx)=0
\qquad \text{a.s.}
\end{align*}
\end{theorem}
%

\medskip

The following corollary establishes the universal consistency of  \proto{}.

\begin{corollary}
\label{thm0}
\textcolor{black}{Let $(\X,\rho)$ be a separable metric space.}
If $m=m_n\to \infty$ such that $m_n/n\to 0$,
then the classification rule $\gproto_{n}$ is universally strongly consistent, that is, for any distribution of $(X,Y)$,
\begin{align*}
\lim_{n\to\infty} L(\gproto_{n})=L^*
\qquad \text{a.s.}
\end{align*}
\end{corollary}


\medskip

\textcolor{black}{The last result of this section is an extension of Theorem~\ref{thm2} 
to real-valued $L^1$-regression 
where the assumption of bounded $Y$ is relaxed to the (necessary) condition $\EXP\{|Y|\}<\infty$. 
We establish consistency for a modified rule, as similarly done in \citet[Theorem 23.3]{GyKoKrWa02}.
Denoting the index of the cell to which $x$ belongs by $\ell(x)\in\{1,\dots,m\}$,
this modified rule is given by
\begin{equation*}
\ft_{n}(x) = 
\begin{cases}
f_n(x) & \text{if } \sum_{i=1}^n \IND_{\{X_i \in A_{m,\ell(x)}\}} \geq \log n,
\\
0 & \text{otherwise.}
\end{cases}
\end{equation*}
}

\begin{theorem}
\textcolor{black}{
\label{thm2_unbounded}
\textcolor{black}{Let $(\X,\rho)$ be a separable metric space.}
If $m=m_n\to \infty$ such that $m_n \log n/n\to 0$,
then the 
estimate $\ft_{n}$ is universally strongly consistent in $L^1$, that is, for any distribution $\dist$ of $(X,Y)$ with $\EXP\{|Y|\}<\infty$,
\begin{align*}
\lim_{n\to\infty} \int |\ft_n(x)-f(x)|\marg(dx)=0
\qquad \text{a.s.}
\end{align*}
}%
\end{theorem}

\textcolor{black}{
By \citet[Theorem 2]{gyorfi1991universal}, Theorem~\ref{thm2_unbounded} implies universal strong consistency of $\ft_n$ also for $L^p$ regression with $1\leq p <\infty$ under the (necessary) condition $\EXP\{|Y|^p\}<\infty$.}

It is interesting to compare
\proto{} to 
the recently proposed
 \optinet{} classifier of \citet{HaKoSaWe20}, which, to date,
was 
the only algorithm known to be universally consistent in any separable metric space.
Denoting the instances in $\D_n$ by $\bX_n = \{X_1,\dots,X_n\}$, \optinet{} first constructs several $\gamma$-nets of $\bX_n$ for different candidate values of $\gamma>0$ 
(a $\gamma$-{net} of $\bX_n$ is any {maximal set} $\bX(\gamma)\subseteq \bX_n$ in which all interpoint distances are at least $\gamma$).
Each $\gamma$-net serves as a nucleus set for a corresponding Voronoi partition of $\X$.
For each such partition, a prototype classifier is constructed by
taking a majority-vote  
in each of its cells.
Then an optimal $\gamma^*$ is selected among the different candidates
by minimizing a compression-based generalization bound over $\gamma$.
Alternatively, 
$\gamma^*$ can be chosen 
via a validation procedure using a hold-out dataset.
\textcolor{black}{The construction time of \optinet{} is $O(n^2)$ and its query time 
depends on the optimal margin chosen at construction.}

\citet{HaKoSaWe20} observed that a model selection procedure for choosing $\gamma^*$ cannot be readily avoided, since one cannot choose {a-priori} a determinstic sequence $\gamma_n$ for which \optinet{} is consistent for all distributions.
This is in contrast to \proto{}, for which any $m = m_n\to\infty$ with $m_n/n\to0$ will do. 
Of course, in practice, $m$ should be chosen based on the data, as done with $\gamma^*$.

\section{Rates of convergence}
\label{sec:rates}
In this section our focus lies on the rate 
at which
the excess error probability 
$$\EXP\{L(g_{n})\}-L^* \to 0.$$
We first derive rates for the $k$-NN rule under the margin condition \eqref{def:marg} and the generalized Lipschitz condition \eqref{def:genLip} (Theorem~\ref{thm:knn_rate}). 
Obtaining convergence rates for the universally consistent \proto{} classifier of Section~\ref{sec:univ} under these conditions (or any other condition mentioned in Section~\ref{sec:intro} for that matter) is currently an open research problem.
Here, we instead derive rates for a second novel prototype rule, 
\hybrid{}, that hybridizes between the \proto{} and $k$-NN rules
(Theorem~\ref{thm:kproto_rate}).
As shown below, \hybrid{} allows a reduction in computational complexity by compressing the data into $m=O(n/k)$ prototypes while enjoying the same rates as the $k$-NN rule.

We first make an additional simplifying assumption.
For a fixed $x \in \X$, let
\begin{equation}\label{defhx}
H_x(r):=\PROB \left(\rho (x,X)\le r\right), \quad r \ge 0,
\end{equation}
be the cumulative distribution function of $\rho (x,X)$. In the sequel, we assume that $H_x(\cdot)$ is continuous for each $x$.
This assumption holds, for example, in the case that $\X = \R^d$ and  $X$ has a 
density.
If $H_x(\cdot)$ is continuous, then in the definition of nearest neighbors, tie happens with probability zero.
In general, one can achieve that $H_x(\cdot)$ is continuous
by adding a randomized component to $X$:
take $\tilde X=(X,U)$ such that $X$ and $U$ are independent, $U$ is uniformly distributed on $[0,1]$, and 
\[
\tilde\rho((x,u),(X,U)):=\rho (x,X)+\delta |u-U|
\]
for some small $\delta>0$.
One can verify that $\tilde\rho$ is indeed a metric.


%
%

\bigskip

\paragraph{Rates for the $k$-NN rule.}
The $k$-nearest neighbor rule is defined as follows.
We fix $x  \in \X$ and reorder the data $(X_1,Y_1),\ldots,(X_n,Y_n)$
according to increasing values of $\rho(x,X_i)$. The reordered data
sequence is denoted by
\[
(X_{(n,1)}(x),Y_{(n,1)}(x)),\ldots,(X_{(n,n)}(x),Y_{(n,n)}(x)).
\]
$X_{(n,k)}(x)$ is the $k$-th nearest neighbor of $x$
where tie breaking is done by indices.
As discussed above, in this paper we assume that tie happens with probability zero.
Choose an integer $k$ less than $n$, then the $k$-nearest-neighbor
estimate of $P_j$ is
\begin{align*}
P_{n,j}(x)=\frac{1}{k}\sum_{i=1}^k \IND_{\{Y_{(n,i)}(x)=j\}},
\end{align*}
and the $k$-nearest-neighbor classification rule is
\begin{equation*}
g_{k,n}(x)=\argmax_{j\in\Y}  P_{n,j}(x).
\end{equation*}
Concerning the properties of the $k$-nearest-neighbor rule   and the related literature see \cite{DeGyLu96}, \cite{GyKoKrWa02}, and \cite{BiDe15}.

In the following theorem we 
bound the rate of convergence of the excess error probability $\EXP\{L(g_{k,n})\}-L^*$ for the $k$-nearest-neighbor classification rule.
In this way we extend the results of
\citet{KoKr07},
\citet{GaKlMa16},
\citet{DoGyWa18} to the multi-class and to the metric space case.
Notice that the paper by
\cite{BiCeGu10} contains rate of convergence results for nearest neighbor regression estimate and  Banach space valued features, which has implications for classification, too;
cf. \cite{ChDa14}.


\begin{theorem}
\label{thm:knn_rate}
\textcolor{black}{Let $(\X,\rho)$ be a separable metric space.}
Assume that the distribution function $H_x(\cdot)$ is continuous for each $x$. If the {\bf \textit{margin condition}} is satisfied 
with $0<\alpha $ 
and the {\bf \textit{generalized Lipschitz condition}} is met,
then
for $k/\log n\to \infty$,
\begin{align*}
\EXP\{L(g_{k,n})\}-L^*
&= O(1/k^{(1+\alpha)/2})+
 O( h(2k/n)^{1+\alpha}).
\end{align*}
\end{theorem}

\medskip

In the case of $\X = \R^d$
and $h(s)= s^{\beta/d}$ as in \eqref{eq:h_Rd},
\begin{align*}
\EXP\{L(g_{k,n})\}-L^*
&= O(1/k^{(1+\alpha)/2})+
 O( (k/n)^{\beta(1+\alpha)/d})
\end{align*}
and the choice
\begin{equation}
\label{kn}
k_n=\lfloor  n^{2\beta/(2\beta + d)}\rfloor
\end{equation}
 yields the order
\begin{equation}
\label{nnratekk}
n^{-\frac{\beta(1+\alpha)}{2\beta + d}}
\end{equation}
as in \eqref{eq:rate_Rd}.
For the two-class problem,
\citet[Theorem 3.5]{AuTs05} showed that, under the strong density assumption and the margin condition with $\alpha\beta\leq d$, (\ref{nnratekk}) is the minimax optimal rate of convergence for the class of 
$\beta$-H\"older-continuous $P_j$'s, that is, the order (\ref{nnratekk}) is a lower bound for {\it any} classifier.

\bigskip
\paragraph{Rates for the \hybrid{} rule.}

We now introduce a second prototype rule, termed \hybrid{}, that hybridizes between  $k$-NN and \proto{}.
\textcolor{black}{It is essentially a private case of the aggregated denoised 1-NN algorithm introduced by \citet{xue2018achieving} and corresponds to the case of using only one subsample. \citet{xue2018achieving} established essentially the optimal rates in \eqref{eq:rate_Rd} for this algorithm in $\R^d$ (with finite-sample guarntees) under the H\"{o}lder condition \eqref{eq:Lip}, the margin condition \eqref{wtsyb}, and the strong density assumption (see Section \ref{sec:intro}).
Here we establish rates under the generalized Lipschitz condition \eqref{gen-Lip} and the margin condition \eqref{wtsyb} in an arbitrary separable metric space.}


\hybrid{} works as follows.
Similarly to \proto{} (see Section~\ref{sec:univ}), an unlabeled sample $\bX'_m$ serves as a nucleus set, inducing a Voronoi partition $\P_m$ of $\X$.
Instead of taking a majority vote in each Voronoi cell,
\hybrid{} stores at each nucleus the majority vote among the $k$-nearest neighbors 
of that nucleus.
Formally, let $X'_{(m,1)}(x)$ be the first nearest neighbor of $x$ among $\bX'_m$.
We fix $x  \in \X$, and let
$X_{(n,k)}(X'_{(m,1)}(x))$ be the $k$-th nearest neighbor of $X'_{(m,1)}(x)$ from the set $\bX_n$ and $Y_{(n,k)}(X'_{(m,1)}(x))$ stands for its label.
The tie breaking is done by randomization. Again, in this section we assume that tie happens with probability $0$.
Choose an integer $k$ less than $n$, then  
\hybrid{}
estimates 
$P_j(x)$ by
the piecewise constant function
\begin{align*}
\hat P_{n,j}(x) = \hat P_{n,j}(X'_{(m,1)}(x))=\frac{1}{k}\sum_{i=1}^k \IND_{\{Y_{(n,i)}(X'_{(m,1)}(x))=j\}},
\end{align*}
and the corresponding prototype nearest-neighbor classification rule is
\begin{align*}
\gkproto_{n}(x)=\argmax_{j\in\Y} 
 \hat P_{n,j}(x).
\end{align*}

\textcolor{black}{The construction and query times of \hybrid{} are $O(mn)$ and $O(m)$ respectively; the same as for \proto{}.
}

\begin{theorem}
\label{thm:kproto_rate}
\textcolor{black}{Let $(\X,\rho)$ be a separable metric space.}
Assume that the distribution function $H_x(\cdot)$ is continuous for each $x$. If the {\bf \textit{margin condition}} is satisfied with $0<\alpha $ and the {\bf \textit{generalized Lipschitz condition}} is met with $h$ that is concave, then
for $k/\log n\to \infty$,
\begin{align*}
\EXP\{L(\gkproto_{n})\}-L^*
&= O(1/k^{(1+\alpha)/2})+
 O( h(k/n)^{1+\alpha})+O( h(1/m)^{1+\alpha}).
\end{align*}
\end{theorem}

According to this theorem, the proper choice of $m$ is proportional to $n/k$.
For $k$ as in \eqref{kn}, we obtain the same optimal rates as in \eqref{nnratekk}.
\textcolor{black}{
Note however that, similarly to the $k$-NN rule, \hybrid{} is not universally consistent in all separable metric spaces, as established by the same counterexample in \citet[Section 3]{cerou2006nearest}. 
}

\textcolor{black}{Lastly, we note that in some cases, the distribution $\mu$ is concentrated on a finite-dimensional subspace of $\X$ such that
on this space the modified Lipschitz condition is satisfied. Then \proto{} and \hybrid{},
not knowing this subspace and the intrinsic dimension, automatically
achieve good rate of convergence; see \citet{kpotufe2012tree} and references therein for other algorithms that are adaptive to the intrinsic dimension.}



\section{Proof of universal consistency}
\label{sec:proofs}

\paragraph{Proof of Theorem~\ref{thm2}.}
Put
\begin{align*}
\bar f_{m}(x)=\frac{\int_{A_{m,\ell}}f(z)\marg(dz)}{\marg( A_{m,\ell})}, 
\quad\mbox{ if } x\in A_{m,\ell}.
\end{align*}
Then,
\begin{align*}
\int |f_n(x)-\bar f_{m}(x)|\marg(dx)
\qquad \text{and} \qquad
\int |\bar f_{m}(x)-f(x)|\marg(dx)
\end{align*}
are called estimation errror and approximation error, respectively.
Introduce the notations
\begin{align*}
\vartheta_n(A)=\frac 1n \sum_{i=1}^n Y_i\IND_{\{X_i\in A\}}, \qquad \vartheta(A)= \EXP\{\vartheta_n(A)\}
\end{align*}
and
\begin{align*}
\marg_n(A)=\frac 1n \sum_{i=1}^n \IND_{\{X_i\in A\}}, \qquad \marg(A)= \EXP\{\marg_n(A)\}.
\end{align*}
Concerning the estimation error,
\begin{align*}
\int |f_n(x)-\bar f_{m}(x)|\marg(dx)
&=
\sum_{V\in \P_{m}}\int_V |f_n(x)-\bar f_{m}(x)|\marg(dx)\\
&=
\sum_{V\in \P_{m}} \left|\frac{\vartheta_n(V) }{\marg_n(V)}-\frac{\vartheta(V) }{\marg(V)}\right|\marg(V).
\end{align*}
If $L$ stands for a bound on $|Y|$, then
$|\vartheta_n(V)| \leq L \marg_n(V)$ and
\begin{align*}
&\int |f_n(x)-\bar f_{m}(x)|\marg(dx)\\
&\le
\sum_{V\in \P_{m}} \left|\frac{\vartheta_n(V) }{\marg_n(V)}-\frac{\vartheta_n(V) }{\marg(V)}\right|\marg(V)
+
\sum_{V\in \P_{m}} \left|\frac{\vartheta_n(V) }{\marg(V)}-\frac{\vartheta(V) }{\marg(V)}\right|\marg(V)\\
&\le
L\sum_{V\in \P_{m}} \left|\marg(V)-\marg_n(V)\right|
+
\sum_{V\in \P_{m}} \left|\vartheta_n(V)-\vartheta(V)\right|.
\end{align*}
The first term of the right hand side is a special case of the second one, therefore we bound only the tail distribution of the second term.
Let $\sigma(\P_{m})$ be the $\sigma$-algebra generated by $\P_{m}$.
As in \cite{BiGy05}, the equality $|a-b|=2(a-b)^+ + (b-a)$ implies
\begin{align*}
\sum_{V\in \P_{m}} \left|\vartheta_n(V)-\vartheta(V)\right|
&=
2\max_{V\in \sigma(\P_{m})} \left(\vartheta_n(V)-\vartheta(V)\right)^+
+
\sum_{V\in\P_m} (\vartheta(V) -  \vartheta_n(V))\\
&=
2\max_{V\in \sigma(\P_{m})} \left(\vartheta_n(V)-\vartheta(V)\right)^+
+
\EXP\{Y\}-\frac 1n \sum_{i=1}^n Y_i.
\end{align*}
Thus, for any $\varepsilon>0$, the Hoeffding inequality implies
\begin{align*}
&\PROB\left\{\sum_{V\in \P_{m}} \left|\vartheta_n(V)-\vartheta(V)\right|\ge \varepsilon\mid \bX'_m\right\}\\
&\le
\PROB\left\{2\max_{V\in \sigma(\P_{m})} \left(\vartheta_n(V)-\vartheta(V)\right)\ge 2\varepsilon/3\mid \bX'_m\right \}
+
\PROB\left\{ \EXP\{Y\}-\frac 1n \sum_{i=1}^n Y_i\ge \varepsilon/3\right \}
\\
&\le
(2^{m}+1)e^{-n\varepsilon^2/(18L^2)}.
\end{align*}
Therefore, $m_n/n\to 0$ together with the Borel-Cantelli lemma implies that
\begin{align*}
\sum_{V\in \P_{m}} \left|\vartheta_n(V)-\vartheta(V)\right|
\to 0
\qquad a.s.
\end{align*}
and the consistency of the estimation error is proved.

Concerning the approximation error, refer to Lemma A.1 in \cite{HaKoSaWe20} such that choose a Lipschitz function $f^*$
with a Lipschitz constant $C$ and with a support contained in a sphere $S$ such that
\begin{align*}
\int | f(x)-  f^*(x)|\marg(dx)
&\le \varepsilon.
\end{align*}
\textcolor{black}{By examining the proof of Lemma A.1 in \cite{HaKoSaWe20}, we may choose $f^*$ to be bounded by the same bound assumed on $|Y|$.}
Put
\begin{align*}
\bar f^*_{m}(x)=\frac{\int_{A_{m,\ell}}f^*(z)\marg(dz)}{\marg( A_{m,\ell})}, 
\quad\mbox{ if } x\in A_{m,\ell}.
\end{align*}
Then,
\begin{align*}
&\int |\bar f_{m}(x)-  f(x)|\marg(dx)\\
&\le
\int |\bar f_{m}(x)-  \bar f^*_{m}(x)|\marg(dx)
+
\int |\bar f^*_{m}(x)-  f^*(x)|\marg(dx)
+
\int |f^*(x)-  f(x)|\marg(dx)\\
&\le
\int |\bar f^*_{m}(x)-  f^*(x)|\marg(dx)
+
2\int |f^*(x)-  f(x)|\marg(dx)\\
&\le
\int |\bar f^*_{m}(x)-  f^*(x)|\marg(dx)
+
2\varepsilon.
\end{align*}
The bound $|Y|\le L$ implies that $|f^*(x)|\le L$, therefore
\begin{align*}
&\int |\bar f^*_{m}(x)-  f^*(x)|\marg(dx)\\
&=
\sum_{V\in \P_{m}}\int_V |\bar f^*_{m}(x)-  f^*(x)|\marg(dx)\\
&=
\sum_{V\in \P_{m}}
\frac{1}{\marg(V)}\int_V \left|\int_V{f^*(z)\marg(dz)}-  f^*(x)\marg(V)\right|\marg(dx)\\
&\le
\sum_{V\in \P_{m}}
\frac{1}{\marg(V)}\int_V \int_V\left|f^*(z)-  f^*(x)\right|\marg(dx)\marg(dz)\\
&\le
\sum_{V\in \P_{m}}
\frac{1}{\marg(V)}\int_V \int_V
\min\{C\rho(x,z),2L\}
\marg(dx)\marg(dz).
\end{align*}
Recall that  $X'_{(m,1)}(x)$ denotes the first nearest neighbor of $x$ among $\bX'_m$.
For $x,z\in V$,
\begin{align*}
\rho(x,z)
&\leq \rho(x,X'_{(m,1)}(x)) + \rho(X'_{(m,1)}(x), z)
=\rho(x,X'_{(m,1)}(x)) + \rho(z, X'_{(m,1)}(z)),
\end{align*}
where we applied that $V$ is a Voronoi cell and so for any  $x,z\in V$, the nucleuses  $X'_{(m,1)}(x)$ and $X'_{(m,1)}(z)$ are identical.
Thus,
\begin{align*}
\int |\bar f^*_{m}(x)-  f^*(x)|\marg(dx)
&\leq
\int \min\{2C\rho(x,X'_{(m,1)}(x)),2L\}\marg(dx).
\end{align*}
\citet{CoHa67} proved that, for a separable metric space,
\begin{align*}
\rho(x,X'_{(m,1)}(x))
\to 0
\end{align*}
a.s.\ as $m\to \infty$, for $\marg$-almost all $x$, cf. Lemma 6.1 in \cite{GyKoKrWa02}.
Therefore, the dominated convergence yields
\begin{align*}
\int |\bar f^*_{m}(x)-  f^*(x)|\marg(dx)\to 0
\qquad a.s.,
\end{align*}
concluding the proof of Theorem~\ref{thm2}.


\medskip
\paragraph{Proof of Corollary~\ref{thm0}.}
An extension of \citet[Theorem 2.2]{DeGyLu96} yields
\begin{align*}
L(\gproto_{n})-L^*
&\le
\sum_{j=1}^M \int |P_j(x)-\tilde P_{n,j}(x)|\marg(dx).
\end{align*}
Thus, the corollary is proved if
\begin{align*}
 \int |P_j(x)-\tilde P_{n,j}(x)|\marg(dx)\to 0
\end{align*}
a.s., $j=1,\dots ,M$, 
which follows from Theorem~\ref{thm2}.

\medskip

\paragraph{\textcolor{black}{Proof of Theorem \ref{thm2_unbounded}}} 
We write the regression estimator $\ft_n$ as
\begin{align*}
\ft_n(x) = \sum_{i=1}^n W_{n,i}(x) Y_i
\end{align*}
where
\begin{align*}
W_{n,i}(x)   
= \frac{\IND_{\{X_i\in A_{m,\ell(x)}\}}}{\sum_{i=1}^n \IND_{\{X_i\in A_{m,\ell(x)}\}}}
\cdot \IND_{\{\sum_{i=1}^n \IND_{\{X_i\in A_{m,\ell(x)}\}} \ge \log n\}}
.
\end{align*}
Note that the weights are sub-probabilities, namely, 
for all $x\in\X$,
$$0\leq \sum_{i=1}^n W_{n,i}(x) \leq 1.$$
By \citet[Theorem 2]{gyorfi1991universal}
(see also \citet[Lemma 23.3]{GyKoKrWa02}),
it suffices to show:
\begin{itemize}
	\item[\textit{(i)}] $\ft_n$ is strongly consistent for $L^1$ assuming $Y$ is bounded, $|Y|\leq L$;
  \item[\textit{(ii)}] there exists $c>0$ such that for any $Y$ with $\EXP\{|Y|\}<\infty$,
\begin{equation}
\label{eq:weights_cond}
\limsup_n \sum_{i=1}^n \int W_{n,i}(x)\marg(dx) |Y_i| \leq c \EXP\{|Y|\}
\quad a.s.
\end{equation}
\end{itemize}
To show \textit{(i)}, assume $|Y|\leq L$ and decompose
\begin{align*}
\int|\ft_n(x) - f(x)|\marg(dx) \leq \int|\ft_n(x) - f_n(x)|\marg(dx) + \int|f_n(x) - f(x)|\marg(dx),
\end{align*}
where $f_n$ is as in \eqref{eq:reg_est}.
By Theorem \ref{thm2},
\begin{align*}
\lim_{n\to\infty} \int |f_{n}(x) - f(x)|\marg(dx)=0
\qquad \text{a.s.}
\end{align*}
Recall the notation $\mu_n(A) = \frac{1}{n}\sum_{i=1}^n \IND_{\{X_i\in A\}}$
for $A\subseteq \X$
and 
let
\begin{align*}
G_n = \{\ell : \mu_n(A_{m,\ell}) \geq \log n /n \}
\qquad\text{and}\qquad
\G_n = \bigcup_{\ell \in G_n } A_{m,\ell}.
\end{align*}
Since $\ft_{n}(x)=f_{n}(x)$ for $x\in\G_n$ and $\ft_{n}(x)=0$ for $x\in\G_n^c$,
\begin{align*}
\int |\ft_{n}(x)-f_{n}(x)|\marg(dx)
 & = \int_{\G_n^c} |\ft_{n}(x)-f_{n}(x)|\marg(dx)
 = \int_{\G_n^c} |f_{n}(x)|\marg(dx)
 \leq L \mu(\G_n^c).
\end{align*}
For $c=e^2$ let
\begin{align*}
F_n = \{\ell  : \mu(A_{m,\ell}) \geq c\log n /n \}
\qquad\text{and}\qquad
\F_n = \bigcup_{\ell \in F_n} A_{m,\ell}.
\end{align*}
Then 
\begin{align*}
\mu(\G_n^c) = \mu( \G_n^c\cap \F_n^c) + \mu(\G_n^c \cap \F_n)
 \leq \mu(\F_n^c) + \mu(\G_n^c \cap \F_n) 
\leq
\frac{c m_n \log n}{n} + \mu(\G_n^c \cap \F_n)
.
\end{align*}
The first term convergences to $0$ by the Theorem's condition on $m_n$.
For the second term,
\begin{align*}
 \mu(\G_n^c \cap \F_n)
&=
\sum_{\ell\in F_n}\mu(A_{m,\ell})\IND_{\{\mu_n(A_{m,\ell})< \log n /n\}} 
\leq
\sum_{\ell\in F_n}\mu(A_{m,\ell})\IND_{\{\mu_n(A_{m,\ell})< \mu(A_{m,\ell})/c\}} 
.
\end{align*}
Chernoff's bound implies that for any $\ell \in F_n$,
\begin{align*}
\PROB\left\{\mu_n(A_{m,\ell})< \mu(A_{m,\ell})/c \mid \bX'_m\right\}
& \leq 
e^{-n \mu(A_{m,\ell}) (1- \frac{1}{c} - \frac{\log c}{c})}
 \leq
e^{-  c(1- \frac{1}{c} - \frac{\log c}{c})\log n}
 \leq
n^{-e^2 + 3}.
\end{align*}
Thus,
for any $0<\eps<1$,
\begin{align*}
&\PROB\left\{\sum_{\ell\in F_n}\mu(A_{m,\ell})\IND_{\{\mu_n(A_{m,\ell})< \mu(A_{m,\ell})/c\}} > \eps\right\}
\\
&=
\EXP\left\{ \PROB\left\{\sum_{\ell\in F_n}\mu(A_{m,\ell})\IND_{\{\mu_n(A_{m,\ell})< \mu(A_{m,\ell})/c\}} > \eps \mid \bX'_m\right\} \right\}
\\
& \leq 
\EXP\left\{ \PROB\left\{\sum_{\ell\in F_n}\mu(A_{m,\ell})\IND_{\{\mu_n(A_{m,\ell})< \mu(A_{m,\ell})/c\}} > \eps \sum_{\ell\in F_n}\mu(A_{m,\ell})\mid \bX'_m\right\} \right\}
\\
& \leq 
\EXP\left\{\sum_{\ell\in F_n} \PROB\left\{\IND_{\{\mu_n(A_{m,\ell})< \mu(A_{m,\ell})/c\}} > \eps \mid \bX'_m\right\} \right\}
\\
&  =
\EXP\left\{\sum_{\ell\in F_n} \PROB\left\{\mu_n(A_{m,\ell})< \mu(A_{m,\ell})/c \mid \bX'_m\right\} \right\}
\\
&\leq 
m_n \cdot n^{-e^2 + 3} 
\leq n^{-e^2 + 4},
\end{align*}
which is summable.
Hence, by the Borel-Cantelli Lemma,
\begin{align*}
\limsup_n\int |\ft_{n}(x)-f_{n}(x)|\marg(dx)  =0 
\quad a.s.,
\end{align*}
concluding the proof of \textit{(i)}.

To show \textit{(ii)}, assume $Y$ satisfies $\EXP\{|Y|\}<\infty$.
We bound
\begin{align*}
\limsup_n \sum_{i=1}^n \int W_{n,i}(x)\left|Y_i \right|
\marg(dx) 
&\leq 
\limsup_n \left(\frac{1}{n}\sum_{i=1}^n |Y_i|\right) \cdot \max_{i}\int n W_{n,i}(x)
\marg(dx).
\end{align*}
By the strong law of large numbers,
$$\frac{1}{n}\sum_{i=1}^n |Y_i| \to \EXP\{|Y|\} 
\quad\text{a.s.}$$
Hence, it suffices to show that for $c=e^2$, with probability one,
\begin{align}
\label{eq:W_c}
\limsup_n \max_{i} \int n  W_{n,i}(x) \marg(dx) \leq c.
\end{align}
If $G_n=\emptyset$, then $W_{n,i}(x)=0$ for all $x\in\X$.
Thus, 
\begin{align*}
\max_{i}\int  n W_{n,i}(x) \marg(dx) = 0.
\end{align*}
If $G_n\neq\emptyset$, 
then since $W_{n,i}(x)=0$ for all $x\in\G_n^c$,
\begin{align*}
\max_{i}\int  n W_{n,i}(x) \marg(dx) 
=
\max_{i:\ell(X_i) \in G_n } \frac{\mu(A_{m,\ell(X_i)})}{\mu_n(A_{m,\ell(X_i)})}.
\end{align*}
Then,
\begin{align*}
\PROB\left\{\max_{i}\int  n W_{n,i}(x) \marg(dx) > c\right\}
&= 
\PROB\left\{G_n\neq\emptyset, \max_{i:\ell(X_i) \in G_n } \frac{\mu(A_{m,\ell(X_i)})}{\mu_n(A_{m,\ell(X_i)})} > c\right\}
\\
&\leq
n \cdot\PROB\left\{\ell(X_1)\in G_n, \frac{\mu(A_{m,\ell(X_1)})}{\mu_n(A_{m,\ell(X_1)})} > c\right\}
\\
& = 
n \cdot\PROB\left\{\ell(X_1)\in G_n\cap F_n, \frac{\mu(A_{m,\ell(X_1)})}{\mu_n(A_{m,\ell(X_1)})} > c\right\}
\\
& \leq 
n \cdot
\EXP\left\{\sum_{\ell\in F_n}\PROB\left\{
X_1\in A_{m,\ell}, \frac{\mu(A_{m,\ell})}{\mu_n(A_{m,\ell})} > c \mid \bX'_m\right\}
\right\}
\\
& =
n \cdot
\EXP\left\{\sum_{\ell\in F_n}\PROB\left\{
X_1\in A_{m,\ell}, \frac{n\mu(A_{m,\ell})}{1 + \sum_{i=2}^n \IND_{\{X_i \in A_{m,\ell}\}}} > c \mid \bX'_m\right\}
\right\}
\\
& =
n \cdot
\EXP\left\{\sum_{\ell\in F_n}
\mu(A_{m,\ell})\cdot
\PROB\left\{
\frac{n\mu(A_{m,\ell})}{1 + \sum_{i=2}^n \IND_{\{X_i \in A_{m,\ell}\}}} > c \mid \bX'_m\right\}
\right\}
\\
& \leq
n \cdot
\EXP\left\{\sum_{\ell\in F_n}
\mu(A_{m,\ell})\cdot
\PROB\left\{\frac{\mu(A_{m,\ell})}{\mu_n(A_{m,\ell})} > c \mid \bX'_m\right\}
\right\}.
\end{align*}
Chernoff's bound implies that for any $\ell \in F_n$,
\begin{align*}
\PROB\left\{\frac{\mu(A_{m,\ell})}{\mu_n(A_{m,\ell})} > c \mid \bX'_m\right\}
& \leq 
e^{-n \mu(A_{m,\ell}) (1- \frac{1}{c} - \frac{\log c}{c})}
 \leq
e^{-(c - 1 - \log c)\log n}
 \leq
n^{-e^2 + 3}.
\end{align*}
Thus,
\begin{align*}
\PROB\left\{\max_{i}\int  n W_{n,i}(x) \marg(dx) > c\right\}
\leq
n^{-e^2 + 4} \cdot
\EXP\left\{\sum_{\ell\in F_n}
\mu(A_{m,\ell}) 
\right\}
\leq n^{-e^2 + 4}
,
\end{align*}
which is again summable. 
Hence, by the Borel-Cantelli Lemma, \eqref{eq:W_c} holds with probability one, concluding the proof of \textit{(ii)} and Theorem~\ref{thm2_unbounded}.

\section{Proofs of convergence rates}
The rates of convergence for the 
$k$-NN and \hybrid{} 
classifiers
of Section~\ref{sec:rates}
are derived using the following decomposition of the excess error probability.

\begin{lemma}
\label{lem:decomposition}
Let $g_n$ be a plug-in rule with estimates $P_{n,j}$ of $P_j$ as in Section~\ref{sec:intro}.
Abbreviating 
\begin{equation*}
\dP_{l}^*(x) = P_{g^*(x)}(x)-P_{l}(x)
\geq 0,
\qquad  l \in \Y,
\end{equation*}
we have
\begin{align*}
\EXP\{L(g_{n})\}-L^*
&\le
\sum_{j=1}^M\sum_{l =1}^M J_{n,j,l}
\end{align*}
where
\begin{equation}
\label{eq:term}
J_{n,j,l} = \int
\dP_{l}^*(x)
\IND_{\{l \ne g^*(x)\}}
\PROB\{|P_{n,j}(x)-P_{j}(x)|\ge \dP_{l}^*(x)/M\}\marg(dx),
\qquad j,l\in\Y.
\end{equation}
\end{lemma}

Below, we bound $J_{n,j,l}$ for each algorithm separately.

\medskip
\paragraph{Proof of Lemma~\ref{lem:decomposition}.}
For any decision function $g$,
\begin{align*}
\PROB\{g(X)\ne Y\mid X\}
&=
1-\PROB\{g(X)= Y\mid X\}\\
&=
1-\sum_{j=1}^M\PROB\{g(X)= Y=j\mid X\}\\
&=
1-\sum_{j=1}^M\IND_{\{g(X)=j\}} P_j(X)\\
&=
1-P_{g(X)}(X),
\end{align*}
which implies
\begin{align*}
\EXP\{L(g_{n})\}-L^*
&=
\EXP\{P_{g^*(X)}(X)-P_{g_n(X)}(X)\}\\
&=
\int \EXP\{(P_{g^*(x)}(x)-P_{g_n(x)}(x))\IND_{\{g^*(x)\ne  g_n(x)\}}\}\marg(dx)\\
&=
\int \EXP\{I_n(x)\}\marg(dx),
\end{align*}
where
\begin{align*}
I_n(x)
&=
(P_{g^*(x)}(x)-P_{g_n(x)}(x))\IND_{\{P_{g^*(x)}(x)> P_{g_n(x)}(x)\}}\IND_{\{ P_{n,g_n(x)}(x)\ge P_{n,g^*(x)}(x)\}}.
\end{align*}
The relation
\begin{align*}
&\{ P_{n,g_n(x)}(x)- P_{n,g^*(x)}(x)\ge 0\}\\
&=
\{ P_{n,g_n(x)}(x)-P_{g_n(x)}(x)+P_{g_n(x)}(x)-P_{g^*(x)}(x)+P_{g^*(x)}(x)-P_{n,g^*(x)}(x)\ge 0\}\\
&\subseteq
\left\{ \sum_{j=1}^M|P_{n,j}(x)-P_{j}(x)|\ge P_{g^*(x)}(x)- P_{g_n(x)}(x)\right\}
\end{align*}
yields
\begin{align*}
&\IND_{\{P_{g^*(x)}(x)> P_{g_n(x)}(x)\}}\IND_{\{ P_{n,g_n(x)}(x)\ge P_{n,g^*(x)}(x)\}}
\le
\IND_{\{ \sum_{j=1}^M|P_{n,j}(x)-P_{j}(x)|\ge P_{g^*(x)}(x)-P_{g_n(x)}(x)>0\}}.
\end{align*}
Thus,
\begin{align*}
&I_n(x)
\le
\sum_{l =1}^M(P_{g^*(x)}(x)-P_{l}(x))\IND_{\{ \sum_{j=1}^M|P_{n,j}(x)-P_{j}(x)|\ge P_{g^*(x)}(x)-P_{l}(x)\}}\IND_{\{ g_n(x)=l \ne g^*(x)\}}\\
&\le
\sum_{l =1}^M(P_{g^*(x)}(x)-P_{l}(x))\IND_{\{ \sum_{j=1}^M|P_{n,j}(x)-P_{j}(x)|\ge P_{g^*(x)}(x)-P_{l}(x)\}}\IND_{\{l \ne g^*(x)\}}\\
&\le
\sum_{j=1}^M\sum_{l =1}^M(P_{g^*(x)}(x)-P_{l}(x))\IND_{\{l \ne g^*(x)\}}\IND_{\{ |P_{n,j}(x)-P_{j}(x)|\ge (P_{g^*(x)}(x)-P_{l}(x))/M\}}
\\&=
\sum_{j=1}^M\sum_{l =1}^M
\dP_{l}^*(x)
\IND_{\{l \ne g^*(x)\}}
\IND_{\{|P_{n,j}(x)-P_{j}(x)|\ge \dP_{l}^*(x)/M\}}.
\end{align*}
Taking expectation and integrating with respect to $\marg$ concludes the proof of Lemma \ref{lem:decomposition}.

\paragraph{Proof of Theorem~\ref{thm:knn_rate}.}
Lemma \ref{lem:decomposition} shows that to bound $\EXP\{L(g_{k,n})\}-L^*$, it suffices to bound $J_{n,j,l}$ in \eqref{eq:term}.
To this end, let
\begin{align*}
\bar P_{n,j}(x)
=\frac{1}{k}\sum_{i=1}^k \EXP\{\IND_{\{Y_{(n,i)}(x)=j\}}\mid X_1,\dots ,X_n\}
=\frac{1}{k}\sum_{i=1}^k P_j(X_{(n,i)}(x)).
\end{align*}
We bound \eqref{eq:term} by
\begin{align*}
J_{n,j,l}
\leq J_{n,j,l}^{(1)} + J_{n,j,l}^{(2)},
\end{align*}
where
\begin{align*}
J_{n,j,l}^{(1)}
&=\int \dP_l^*(x)\IND_{\{l \ne g^*(x)\}}
\PROB\{|P_{n,j}(x)-\bar P_{n,j}(x)|\ge \dP_l^*(x)/(2M)\}\marg(dx)
\end{align*}
and
\begin{align*}
J_{n,j,l}^{(2)}
&=\int \dP_l^*(x)\IND_{\{l \ne g^*(x)\}}
\PROB\{|\bar P_{n,j}(x)-P_{j}(x)|\ge \dP_l^*(x)/(2M)\}\marg(dx).
\end{align*}
For each $j$ and $l$, we show that
\begin{align*}
J_{n,j,l}^{(1)}
&=
O(1/k^{(1+\alpha)/2})
\end{align*}
and
\begin{align*}
J_{n,j,l}^{(2)}
&=
O( h(2k/n)^{1+\alpha}),
\end{align*}
from which the theorem follows.

The estimation error $J_{n,j,l}^{(1)}$ can be managed in the same way as in the proof of Lemma 1 in \cite{DoGyWa18}.
The Hoeffding inequality implies that
\begin{align*}
&\PROB\{|P_{n,j}(x)-\bar P_{n,j}(x)|\ge  \dP_l^*(x)/(2M)\mid X_1,\dots ,X_n\}
\le
2e^{-k\dP_l^*(x)^2/(2M^2) }.
\end{align*}
Therefore,
\begin{align*}
J_{n,j,l}^{(1)}
&\le
2\EXP\left\{ \dP_l^*(X)\IND_{\{l \ne g^*(X)\}}  e^{-k\dP_l^*(X)^2/(2M^2) } \right\}.
\end{align*}
The margin condition with parameter $\alpha$ means that
for $0\le t\le 1$,
\begin{align*}
G(t)
:=
\PROB\{\dP_l^*(X)\IND_{\{l \ne g^*(X)\}}\le t\}
&\le
\PROB\{P_{(1)}(X)-P_{(2)}(X)\le t\}\le c^* t^{\alpha}
.
\end{align*}
This implies that
\begin{align}
\label{ka}
J_{n,j,l}^{(1)}
&\le
2\int_0^1 s e^{-ks^2/(2M^2)}G(ds)\nonumber\\
&\le
2c^*\alpha\int_0^1 s e^{-ks^2/(2M^2)}s^{\alpha-1}ds\nonumber\\
&\le
2c^*\alpha
\cdot
k^{-(\alpha+1)/2}\cdot
\int_0^{\infty} e^{-u^2/(2M^2)}
u^{\alpha}du
\nonumber
\\
&=
O( k^{-(\alpha+1)/2}).
\end{align}

Concerning the approximation error $J_{n,j,l}^{(2)}$, we follow the line of proof of Lemma 2 in \cite{DoGyWa18}.
The generalized Lipschitz condition implies that
\begin{align*}
|\bar P_{n,j}(x)-P_{j}(x)|
&\le
\frac{1}{k}\sum_{i=1}^k |P_j(X_{(n,i)}(x))- P_j(x)|\\
&\le
\frac{1}{k}\sum_{i=1}^k h(\marg(S_{x,\rho(x,X_{(n,i)}(x))}))\\
&\le
h(\marg(S_{x,\rho(x,X_{(n,k)}(x))})).
\end{align*}
If the distribution function $H_x(\cdot)$ is continuous for each $x$, then
we apply the probability integral transform
(cf.
\cite{BiDe15}, p.\ 8).
As a result, the random variable
\begin{align}
\label{HH}
H_x(\rho (x,X))=\marg \{ S_{x,\rho (x,X)} \}
\end{align}
is uniformly distributed on $[0,1]$.
For i.i.d.\ uniformly distributed $U_1,\dots ,U_n$, 
denote by $U_{(1,n)},\dots ,U_{(n,n)}$ the corresponding order statistics.
Then (\ref{HH}) implies
\begin{align*}
\marg ( S_{x,\rho(x,X_{(n,k)}(x))})
\eD U_{(k,n)}.
\end{align*}
Thus, from the generalized Lipschitz condition one gets
\begin{align}
\PROB\{  \dP_l^*(x)/(2M)\le |\bar P_{n,j}(x)-P_{j}(x)|\}\nonumber
&\le
\PROB\left\{  \dP_l^*(x)/(2M)\le  h(\marg(S_{x,\rho(x,X_{(n,k)}(x))}))\right\}\nonumber\\
&=
\PROB\left\{  \dP_l^*(x)/(2M)\le  h(U_{(k,n)})\right\}\nonumber\\
&=
\PROB\left\{ h^{-1}( \dP_l^*(x)/(2M))\le U_{(k,n)}\right\}.
\label{*}
\end{align}
As in the proof of Lemma 3
of \cite{DoGyWa18}, Chernoff's exponential inequality implies
\begin{align}
&\PROB\{  \dP_l^*(x)/(2M)\le |\bar P_{n,j}(x)-P_{j}(x)|\}\nonumber\\
&\le
\PROB\left\{ \sum_{i=1}^n\IND_{\{ U_i\le h^{-1}( \dP_l^*(x)/(2M))\}}< k\right\}\nonumber\\
&\le
e^{-(1-\log 2)k }
+\IND_{\{ h^{-1}( \dP_l^*(x)/(2M))< 2k/n\}}.
\label{**}
\end{align}
Applying the margin condition, we get
\begin{align*}
J_{n,j,l}^{(2)}
&\le
\EXP\Big\{  \dP_l^*(X)\IND_{\{l \ne g^*(X)\}}
(e^{-(1-\log 2)k }+\IND_{\{ h^{-1}( \dP_l^*(X)/(2M))< 2k/n\}})\Big\}\\
&\le e^{-(1-\log 2)k } +\int_0^1 s \IND_{\{ h^{-1}(s/(2M))< 2k/n\}}G(ds)\\
&\le e^{-(1-\log 2)k } +c^*\alpha\int_0^1 s^{\alpha} \IND_{\{ s< 2Mh(2k/n)\}}ds\\
&\le e^{-(1-\log 2)k } +
O( h(2k/n)^{1+\alpha}),
\end{align*}
concluding the proof of Theorem~\ref{thm:knn_rate}.


%
%
%
%

\paragraph{Proof of Theorem~\ref{thm:kproto_rate}.}
As in the proof of Theorem~\ref{thm:knn_rate}, we 
bound \eqref{eq:term} by
\begin{align*}
J_{n,j,l}
\leq J_{n,j,l}^{(1)} + J_{n,j,l}^{(2)},
\end{align*}
where
\begin{align*}
J_{n,j,l}^{(1)} &=\int  \dP_l^*(x)\IND_{\{l \ne g^*(x)\}}
\PROB\{|P_{n,j}(X'_{(m,1)}(x))-\bar P_{n,j}(X'_{(m,1)}(x))|>  \dP_l^*(x)/(2M)\}\marg(dx)
\end{align*}
and
\begin{align*}
J_{n,j,\ell}^{(2)}
&=\int  \dP_l^*(x)\IND_{\{l \ne g^*(x)\}}
\PROB\{|\bar P_{n,j}(X'_{(m,1)}(x))-P_{j}(x)|>  \dP_l^*(x)/(2M) \}\marg(dx).
\end{align*}
For each $j$ and $l$, we show that
\begin{align*}
J_{n,j,l}^{(1)}
&=
O(1/k^{(1+\alpha)/2})
\end{align*}
and
\begin{align*}
J_{n,j,l}^{(2)}
&=
O( h(k/n)^{\alpha+1})+
O( h(1/m)^{\alpha+1}),
\end{align*}
from which the theorem follows.
The Hoeffding inequality implies that
\begin{align*}
&\PROB\{|P_{n,j}(X'_{(m,1)}(x))-\bar P_{n,j}(X'_{(m,1)}(x))|\ge  \dP_l^*(x)/(2M)
 \mid 
\bX_n
 ,X'_{(m,1)}(x)\}
\le
2e^{-k \dP_l^*(x)^2/(2M^2) }.
\end{align*}
Therefore, similarly to (\ref{ka}), the margin condition implies  that
\begin{align*}
J_{n,j,l}^{(1)}
&\le
2\EXP\left\{  \dP_l^*(X)\IND_{\{l \ne g^*(X)\}}  e^{-k \dP_l^*(X)^2/(2M^2) } \right\}
=
O( k^{-(\alpha+1)/2}).
\end{align*}
The generalized Lipschitz condition implies that
\begin{align*}
&|\bar P_{n,j}(X'_{(m,1)}(x))-P_{j}(x)|\\
&\le
\frac{1}{k}\sum_{i=1}^k |P_j(X_{(n,i)}(X'_{(m,1)}(x)))- P_j(X'_{(m,1)}(x))|+|P_j(X'_{(m,1)}(x))-P_j(x)|\\
&\le
\frac{1}{k}\sum_{i=1}^k h(\marg(S_{X'_{(m,1)}(x),\rho(X'_{(m,1)}(x),X_{(n,i)}(X'_{(m,1)}(x)))})) +h(\marg(S_{x,\rho(x,X'_{(m,1)}(x))}))\\
&\le
h(\marg(S_{X'_{(m,1)}(x),\rho(X'_{(m,1)}(x),X_{(n,k)}(X'_{(m,1)}(x)))}))+h(\marg(S_{x,\rho(x,X'_{(m,1)}(x))})).
\end{align*}
Thus,
\begin{align*}
&\PROB\{ \dP_l^*(x)/(2M)\le |\bar P_{n,j}(X'_{(m,1)}(x))-P_{j}(x)|\}\\
&\le
\PROB\left\{  \dP_l^*(x)/(4M)\le h(\marg(S_{X'_{(m,1)}(x),\rho(X'_{(m,1)}(x),X_{(n,k)}(X'_{(m,1)}(x)))}))\right\}\\
&\quad+
\PROB\left\{  \dP_l^*(x)/(4M)\le  h(\marg(S_{x,\rho(x,X'_{(m,1)}(x))}))\right\}.
\end{align*}
Similarly to (\ref{*}) and (\ref{**})  in the proof of Theorem \ref{thm:knn_rate}, Chernoff's 
inequality implies
\begin{align*}
&
\PROB\left\{  \dP_l^*(x)/(4M)\le  h(\marg(S_{X'_{(m,1)}(x),\rho(X'_{(m,1)}(x),X_{(n,k)}(X'_{(m,1)}(x)))}))\right\}\\
&=
\PROB\left\{  \dP_l^*(x)/(4M)\le  h(U_{(k, n)})\right\}\\
&=
\PROB\left\{ h^{-1}( \dP_l^*(x)/(4M))\le  U_{(k, n)}\right\}\\
&\le
\PROB\left\{ \sum_{i=1}^{ n}\IND_{\{ U_i\le h^{-1}( \dP_l^*(x)/(4M))\}}< k\right\}\\
&\le
e^{-(1-\log 2)k }
+\IND_{\{ h^{-1}( \dP_l^*(x)/(4M))< 2k/n\}}.
\end{align*}
Applying the margin condition, we get
\begin{align*}
&\EXP\Big\{  \dP_l^*(X)\IND_{\{l \ne g^*(X)\}}
\PROB\{  \dP_l^*(X)/(4M)\le h(\marg(S_{X'_{(m,1)}(X),\rho(X'_{(m,1)}(X),X_{(n,k)}(X'_{(m,1)}(X)))}))\mid X\}\Big\}
\\
&\le
\EXP\Big\{  \dP_l^*(X)\IND_{\{l \ne g^*(X)\}}
(e^{-(1-\log 2)k }+\IND_{\{ h^{-1}(\dP_l^*(X)/(4M))< 2k/ n\}})\Big\}\\
&\le e^{-(1-\log 2)k } +\int_0^1 s \IND_{\{ h^{-1}(s/(4M))< 2k/ n\}}G(ds)\\
&\le e^{-(1-\log 2)k } +c^*\alpha\int_0^1 s^{\alpha} \IND_{\{ s< 4Mh(2k/ n)\}}ds\\
&\le e^{-(1-\log 2)k } + O( h(2k/ n)^{\alpha+1})\\
&= e^{-(1-\log 2)k } + O( h(k/ n)^{\alpha+1}),
\end{align*}
where in the last equality we applied the assumption that $h$ is concave.
A slight modification of the previous argument yields
\begin{align*}
\PROB\left\{  \dP_l^*(x)/(4M)<  h(\marg(S_{x,\rho(x,X'_{(m,1)}(x))}))   \right\}\nonumber
&=
\PROB\left\{  \dP_l^*(x)/(4M)< h(U_{(1, m)})\right\},
\end{align*}
and so
\begin{align*}
&
\EXP\Big\{  \dP_l^*(X)\IND_{\{l \ne g^*(X)\}}
 \PROB\{  \dP_l^*(X)/(4M)<h(\marg(S_{X,\rho(X,X'_{(m,1)}(X))}))\mid X\}\Big\}\\
&=
\EXP\Big\{  \dP_l^*(X)\IND_{\{l \ne g^*(X)\}}
\PROB\left\{  \dP_l^*(X)/(4M)< h(U_{(1, m)})\mid X\right\}\Big\}\\
&\le \EXP\left\{\int_0^1 s \IND_{\{ s< 4Mh(U_{(1, m)})\}}G(ds)\right\}\\
&\le c^*\alpha\EXP\left\{\int_0^1 s^{\alpha} \IND_{\{ s< 4Mh(U_{(1, m)})\}}ds\right\}\\
&=
O( \EXP\left\{h(U_{(1, m)})^{\alpha+1}\right\})\\
&=
O( h(1/m)^{\alpha+1}),
\end{align*}
where in the last equality we again applied the assumption that $h$ is concave.

\vskip 0.2in
\textcolor{black}{
\paragraph{Acknowledgments.} We thank the editor and referees for carefully reading 
the manuscript and for the suggested improvements.
We also thank Roberto Colomboni for pointing out some inaccuracies in a previous version of the manuscript.
}

\vskip 0.2in
\bibliography{refs}

\end{document}